\documentclass{article}
\usepackage{spconf,amsmath,graphicx}
\usepackage{amssymb}
\usepackage{multirow}
\usepackage{colortbl}  
\usepackage{xcolor}
\usepackage{array}
\usepackage{stfloats}

\usepackage[ruled,vlined]{algorithm2e}
\usepackage{bm}
\title{EPA: Neural Collapse Inspired Robust Out-of-Distribution Detector}
\name{Jiawei Zhang\textsuperscript{1}\textsuperscript{*}, Yufan Chen\textsuperscript{2}\textsuperscript{*}, Cheng Jin\textsuperscript{1}, Lei Zhu\textsuperscript{2}, Yuantao Gu\textsuperscript{1}\textsuperscript{\dag}\thanks{\textsuperscript{*}Equal Contribution. \textsuperscript{\dag}Corresponding Author.}\thanks{This work was supported in part by NSAF (Grant No. U2230201) and a Grant from the Guoqiang Institute, Tsinghua University.}}
\address{\textsuperscript{1}
Department of Electronic Engineering, Tsinghua University, Beijing 100084, China\\
\textsuperscript{2}College of Communication Engineering, Army Engineering University of PLA}
\begin{document}
\ninept
\maketitle
\begin{abstract}
% Out-of-Distributi (OOD) detection plays a crucial role in ensuring the security of neural networks. Recent studies have leveraged the fact that In-Distribution (ID) samples form a subspace in the feature space, achieving state-of-the-art (SOTA) performance. However, these methods overlook the fine-grained distribution of ID features within the feature subspace. In this paper, we present innovative insights into the distribution of ID features within the corresponding feature subspace based on Neural Collapse ($\mathcal{NC}$). Building upon these observations, we propose a novel OOD scoring function, named Entropy enhanced Principal Angle (EPA). In comparison to existing methods, our EPA score not only considers the low-dimensional subspace formed by ID features but also accounts for fine-grained characteristics within the subspace. We experimentally compare EPA with various SOTA approaches, validating its superior performance and robustness across different network architectures and OOD datasets.
Out-of-distribution (OOD) detection plays a crucial role in ensuring the security of neural networks. Existing works have leveraged the fact that In-distribution (ID) samples form a subspace in the feature space, achieving state-of-the-art (SOTA) performance. However, the comprehensive characteristics of the ID subspace still leave under-explored. Recently, the discovery of Neural Collapse ($\mathcal{NC}$) sheds light on novel properties of the ID subspace. Leveraging insight from $\mathcal{NC}$, we observe that the Principal Angle between the features and the ID feature subspace forms a superior representation for measuring the likelihood of OOD. Building upon this observation, we propose a novel $\mathcal{NC}$-inspired OOD scoring function, named Entropy-enhanced Principal Angle (EPA), which integrates both the global characteristic of the ID subspace and its inner property. We experimentally compare EPA with various SOTA approaches, validating its superior performance and robustness across different network architectures and OOD datasets.
\end{abstract}
\vspace{-0.1em}
\section{Introduction}
\vspace{-0.1em}
\label{sec:intro}

Neural networks confront security challenges arising from the presence of Out-of-distribution (OOD) samples. A reliable model should not only precisely classify In-Distribution (ID) samples but also effectively discern and reject unknown samples. This highlights the significance of the OOD detection task, which aims to discriminate whether a given sample pertains to the ID category or not.

% Existing methods aim to design scoring functions to distinguish between ID and OOD samples based on potential disparities originating from three sources: softmax probabilities \cite{b1,b2}, logits \cite{b3,b4,b5}, and feature space \cite{b6,b7,b8,b9,b10,b11,b12}. The reduction from feature space to logits may result in the omission of certain feature details and potentially impair the performance of OOD detection \cite{b6}. Thus the majority of recent approaches primarily leverage feature space attributes in the design of scoring functions. It has been observed that ID features form a subspace within the feature space, giving rise to the development of several competitive methods \cite{b6, b7, b8}. However, these works have mainly focused on the characteristic of OOD samples being distant from the ID subspace, without considering the fine-grained distribution within the ID subspace that may assist in identifying more discriminative attributes.

Existing methods aim to design scoring functions to distinguish between ID and OOD samples based on potential disparities originating from three sources: softmax probabilities \cite{b1,b2}, logits \cite{b3,b4,b5}, and feature space \cite{b6,b7,b8,b9,b10,b11,b12}. Among them, the majority of recent approaches primarily leverage feature space attributes in the design of scoring functions. It has been observed that ID features form a subspace within the feature space, giving rise to the development of several competitive methods \cite{b6, b7, b8}. However, these works often utilize the subspace property without delving into a corresponding explanation and more comprehensive characteristics which may assist in identifying more interpretable and discriminative attributes.

Recently, the discovery of the Neural Collapse ($\mathcal{NC}$) phenomenon and its associated theories have revealed four distinct characteristics of ID features during the terminal phase of training (TPT) \cite{b15}. Viewed from a ``bottom-up" perspective, $\mathcal{NC}$ elucidates the prevalent operational paradigms underlying feature engineering in neural networks and propels the advancement of neural network interpretability with effective and novel insights to the subdomains of machine learning \cite{b25,b26}. For instance, researchers have quantitatively observed that $\mathcal{NC}$ consistently occurs during the transportation of a model to novel samples or categories, leading to a predictable indicator of the performance of transfer learning and the development of novel incremental learning methods \cite{b27, b29}. The geometric properties of training features elucidated by $\mathcal{NC}$ motivate us to delve deeper into the comprehensive attributes of the ID subspace for more effective OOD detection methods.

% Drawing inspiration from $\mathcal{NC}$, we shed light on a vital attribute of ID features: as training progresses, the features of training samples gradually depart from the global feature mean and collapse towards the mean of their respective categories, leading to a considerable distance from the ID features to the global feature mean. This insight indicates that ID features have significant components within the corresponding subspace and lie farther from the global feature mean, which highlights that the Principal Angle (PA) between the feature and the ID subspace could be employed as a discriminative indicator for OOD detection. Meanwhile, existing research suggests that the integration of multiple aspects could boost the detection performance \cite{b6, b12}. Thus, we further enhance the performance of PA by integrating widely employed Entropy method \cite{b12, b30, b31} to fuse feature space and softmax information.
Drawing inspiration from $\mathcal{NC}$, we shed light on a vital attribute of ID features: as training progresses, the features of training samples gradually depart from the global feature mean and collapse towards the mean of their respective categories, leading to a considerable distance from the ID features to the global feature mean. This insight indicates that ID features have significant components within the corresponding subspace and lie farther from the global feature mean, which highlights that the \textit{Principal Angle} (PA) between the feature and the ID subspace could be employed as a discriminative indicator for OOD detection. Moreover, while PA characteristics the global property of the subspace, its inner structure obtained from $\mathcal{NC}$ induces a higher concentration of the output probabilities of ID samples, which could be measured through the widely employed Entropy method \cite{b12, b30, b31}. This motivates us to further enhance the performance of PA by integrating softmax Entropy to fuse the global and the inner information of the subspace.
% Through comprehensive experimental comparisons against state-of-the-art (SOTA) methods, we have observed that the $\mathcal{NC}$-inspired algorithm proposed in this paper not only outperforms existing methods but also demonstrates greater robustness.

% The distinctive attributes of training samples obtained from $\mathcal{NC}$ offer crucial cues for distinguishing between ID and OOD samples, rendering 
The application of $\mathcal{NC}$ in OOD detection tasks is a novel and promising perspective. To the best of our knowledge, only one relevant study currently enhances the performance of OOD detection by inducing $\mathcal{NC}$ using L2 regularization on features during the training process \cite{b16}. Nevertheless, non-post-hoc methods may have limited applicability to pre-trained models, and a corresponding explanation of the association between $\mathcal{NC}$ and OOD detection has yet to be explored. Furthermore, the emergence of $\mathcal{NC}$ often coincides with the enhancement of generalization capability and robustness \cite{b15}. This propels us to further explore whether the $\mathcal{NC}$-inspired OOD detection method exhibits superior robustness compared to existing methods, since real-world applications request consistent effectiveness across various models and datasets while reliably addressing tasks of varying difficulties. \par

In this paper, we provide novel insights to bridge the gap in the relationship between $\mathcal{NC}$ and OOD detection. Building upon this exploration, we propose an $\mathcal{NC}$-inspired OOD scoring function named \textit{\textbf{E}ntropy 
 enhanced \textbf{P}rincipal \textbf{A}ngle} (EPA), which not only considers the overarching subspace structure of ID features but also encompasses the inner property within the subspace. We validate the outstanding performance and robustness of EPA by comprehensive comparison with various state-of-the-art (SOTA) methods on diverse network architectures and datasets, demonstrating the potential of reconsidering the OOD detection problem from the perspective of $\mathcal{NC}$ for future research.
\section{Preliminaries}
\label{sec:format}

In this section, we present the notations and fundamental concepts of OOD detection and Neural Collapse, laying the necessary groundwork for our approach.
\subsection{Notations}
Denote the training distribution as $\mathbb{X}_{\text{id}}\times\mathbb{Y}$, where $\mathbb{X}_{\text{id}}$ represents the data distribution of the training set, and $\mathbb{Y}$ represents the corresponding categories with the number of categories $|\mathbb{Y}|=C$. Given a neural network classifier $\mathbf{f}(\mathbf{x}):\mathbb{X}_{\text{id}}\rightarrow \mathbb{R}^{|\mathbb{Y}|}$ trained on $\mathbb{X}_{\text{id}}\times \mathbb{Y}$, assume $\mathbf{f}(\mathbf{x})=\mathbf{W}^\mathrm{T}\mathbf{g}(\mathbf{x})+\mathbf{b}$, where $\mathbf{z}=\mathbf{g}(\mathbf{x})\in \mathbb{R}^n$ is the penultimate layer feature corresponding to the sample $\mathbf{x}$. $\mathbf{W}=(\mathbf{w}_1, \mathbf{w}_2, \cdots, \mathbf{w}_C)\in \mathbb{R}^{n\times C}$ and $\mathbf{b}\in \mathbb{R}^C$ are the weights and bias of the last linear layer in the network.

Let $\mathbf{x}_{i,c}$ be the $i$th training sample in category $c$, with its corresponding penultimate layer feature as $\mathbf{z}_{i,c}=\mathbf{g}(\mathbf{x}_{i,c})$. The feature mean for category $c$ and the global feature mean are denoted as $\bm{\mu}_c=\mathbb{E}_i(\mathbf{z}_{i,c})$ and $\bm{\mu}_g=\mathbb{E}_{i,c}(\mathbf{z}_{i,c})$ respectively. 
\subsection{OOD Detection}

 The goal of OOD detection is to determine whether a new sample $\mathbf{x}$ belongs to or lies outside of $\mathbb{X}_{\text{id}}$. The mainstream OOD detection methods aims to design a scoring function $h(\mathbf{x})$, which quantifies the likelihood that a sample $\mathbf{x}$ is classified as an OOD sample. 
 Thus, OOD detection can be formulated as a binary classification problem as follows.
\begin{equation}
\label{OOD}
    \begin{aligned}S(\mathbf{x})=\begin{cases}\text{OOD} &h(\mathbf{x})\ge \gamma\\\text{ID} &h(\mathbf{x})<\gamma\end{cases}\end{aligned}
\end{equation}
where $\gamma$ is a threshold. 
Commonly used evaluation metrics include the Area Under the Receiver Operating Characteristic curve (AUROC) and the False Positive Rate when the True Positive Rate reaches 95\% (FPR@95). 
AUROC and FPR@95 values range from 0 to 1. 
A higher AUROC value and a lower FPR@95 value indicate better performance of the scoring function. 

\subsection{Neural Collapse}
 In \cite{b15}, four phenomena related to the behavior of the penultimate layer features of the neural networks are observed during the terminal phase of training:
($\mathcal{NC}1$) The variance of the features within the same category tends to be zero. ($\mathcal{NC}2$) The category means of features converge to a general simplex equiangular tight frame (ETF). ($\mathcal{NC}3$) The difference between the rescaled weight matrix of the last layer and the rescaled category feature mean converges to zero. ($\mathcal{NC}4$) The network simplifies to a nearest class center classifier.

\section{Methodology}
\label{sec:pagestyle}

Neural Collapse reveals both the global and inner structures of the subspace formed by features in a neural network during the late stages of training. While some prior literature \cite{b6,b7,b8} has also considered the use of subspace properties of features, further investigation is still required to delve into a corresponding explanation and more comprehensive properties of the subspace. In this section, we begin by introducing our novel observations based on Neural Collapse and presenting guidelines for designing the scoring functions. Then we provide details on the implementation of the proposed EPA method.

\subsection{Insights from $\mathcal{NC}$}

We draw the following conclusions from $\mathcal{NC}1$ to $\mathcal{NC}4$: When $\mathcal{NC}$ happens, we have

\begin{itemize} 

\item[(a)] The features of training samples (after mean subtraction) form a subspace $\mathbb{V}_{\text{id}}$, and the distance between training features and the origin of $\mathbb{V}_{\text{id}}$ is approximately a constant value $\alpha>0$, where $\alpha$ is the distance between the vertices and the center of the general simplex ETF.

\item[(b)] The difference between $\mathbf{o}^{\prime}=-\left(\mathbf{W}^\mathrm{T}\right)^{\dagger}\mathbf{b}\in \mathbb{R}^n$ and the global feature mean $\bm{\mu}_g$ is approximately orthogonal to the feature subspace $\mathbb{V}_{\text{id}}$. Here, $\left(\mathbf{W}^\mathrm{T}\right)^{\dagger}$ denotes the Moore-Penrose generalized inverse of $\mathbf{W}^\mathrm{T}$.

\item[(c)] The outputs of the training samples exhibit high concentration.

\end{itemize}

Conclusion (a) is derived from $\mathcal{NC}1$ and $\mathcal{NC}2$, as the vertices of the ETF constitute a subspace in $\mathbb{R}^n$. Conclusion (c) originates from the equiangular property of the ETF and the nearest class center classifier. Conclusion (b) comes from $\mathcal{NC}3$ and $\mathcal{NC}4$. Consider 
\begin{equation}
   \begin{aligned}
    \mathbf{y}=\mathbf{W}^\mathrm{T}\mathbf{z}+\mathbf{b}\approx \mathbf{W}^\mathrm{T}\left(\mathbf{z}-\mathbf{o}^{\prime}\right) =\mathbf{W}^\mathrm{T}\left(\mathbf{z}-\bm{\mu}_g+\bm{\mu}_g-\mathbf{o}^{\prime}\right)
\end{aligned} 
\end{equation}
Since $\mathbf{w}_c$ is parallel to $\bm{\mu}_c-\bm{\mu}_g$ and $||\mathbf{w}_c||_2\approx||\mathbf{w}_{c'}||_2$ (which can be obtained from $\mathcal{NC}3$), for any two categories $c,c'$, to ensure the nearest neighbor classifier's characteristics, it is required that $\bm{\mu}_g-\mathbf{o}^\prime$ is orthogonal to $\bm{\mu}_c-\bm{\mu}_{c'}$. By iterating over all $c,c'$, $\bm{\mu}_g-\mathbf{o}^\prime$ needs to be orthogonal to the subspace $\mathbb{V}_{\text{id}}$.

The above observations inspire us to utilize the \textit{Principal Angle} (abbreviated as PA) between the features and the subspace to measure the likelihood of a sample belonging to the ID category. Conclusion (a) indicates that features of ID samples are likely to be in a subspace and far from the origin, so they enjoy small PAs, while features of OOD samples may not possess the same characteristics as ID samples. In practice, biasing the features by $o^\prime$ can introduce favorable improvement in the performance. We explain this by conclusion (b), as when redefining $o^\prime$ as the origin, the distance of ID features from the origin will be enlarged to $d\approx \sqrt{\alpha^2+||\bm \mu_g-o^\prime||^2}$, which leads to smaller PAs. 

PA leverages the global information of the subspace by measuring the affinity between the feature and the subspace, while the inner structure induced from $\mathcal{NC}$ remains underutilized. We note that conclusion (c) actually delves into the equiangular property of the ETF and provides complementary information for PA. This motivates the further enhancement of PA by quantifying the concentration through softmax \textit{Entropy} and fusing information from both perspectives. Below we provide the specific form of the proposed EPA method, which consists of the computation of PA and a training set-dependent integrating procedure with Entropy.

\subsection{The proposed EPA}
\begin{algorithm}[ht]
  \SetAlgoLined
  \KwIn{Training subset $\mathbb{X}_s$, Network weights $\mathbf{g}(\cdot)$, $\mathbf{W}$, $\mathbf{b}$, Hyperparameter $D$, Test sample $\mathbf{x}$.}
  \KwOut{EPA score of $\mathbf{x}$.}
  \caption{The implementation of EPA}
  \label{alg:spa}
  1. Calculate $\mathbf{o}^\prime=-(\mathbf{W}^\mathrm{T})^{\dagger}\mathbf{b}$.
  
  2. Denote the orthogonal decomposition as $\mathbb{E}_{\mathbf{x}_s\in \mathbb{X}_s}\left(\overline{\mathbf{g}}(\mathbf{x}_s)\overline{\mathbf{g}}(\mathbf{x}_s)^T\right)=\mathbf{P}\mathbf{\Lambda} \mathbf{P}^\mathrm{T}$ with $\mathbf{\overline{g}}(\mathbf{x}_s)=\mathbf{{g}}(\mathbf{x}_s)-o^\prime$. Take the first $D$ columns of $\mathbf{P}$ as $\mathbf{R}$.

  3. Calculate the scaling factor 
  $\beta=\frac{\max_{\mathbf{x}_s\in \mathbb{X}_s}H(\mathbf{x}_s)}{\min_{\mathbf{x}_s\in \mathbb{X}_s}\theta(\mathbf{x}_s)}$, where $\theta(\cdot)$ and $H(\cdot)$ are defined in (\ref{eq:theta}) and (\ref{H}).
  
  4. The EPA score of $\mathbf{x}$ is 
  $\text{EPA} (\mathbf{x})=\beta\theta(\mathbf{x})+H(\mathbf{x})$.
  
  \Return $\text{EPA}(\mathbf{x})$
\end{algorithm}
\textit{\textbf{Computation of PA.}} Firstly, we sample a subset $\mathbb{X}_s$ from the training set to compute the subspace. Denote $\overline{\mathbf{g}}(\mathbf{x})=\mathbf{g}(\mathbf{x})-\mathbf{o}^\prime$, and consider the orthogonal decomposition of $\mathbb{E}_{\mathbf{x}_s\in \mathbb{X}_s}\left(\overline{\mathbf{g}}(\mathbf{x}_s)\overline{\mathbf{g}}(\mathbf{x}_s)^T\right)$ as $\mathbf{P}\mathbf{\Lambda} \mathbf{P}^\mathrm{T}$, assuming that $\mathbf{\Lambda}$ is arranged in descending order of eigenvalues. Let $\mathbf{R}$ be the matrix composed of the first $D$ columns of $\mathbf{P}$, then the projection matrix onto the subspace formed by the first $D$ eigenvectors of $\mathbf{\Sigma}_s$ is $\mathbf{Q}=\mathbf{R}\mathbf{R}^\mathrm{T}$.

Given a test sample $\mathbf{x}$, we can calculate the length of the parallel component of $\mathbf{\overline{g}}(\mathbf{x})$ within the subspace, which is given by 
\begin{equation}
    ||\mathbf{Q}\overline{\mathbf{g}}(\mathbf{x})||_2=||\mathbf{R}^\mathrm{T}\overline{\mathbf{g}}(\mathbf{x})||_2
\end{equation}
Hence, the PA between $\overline{\mathbf{g}}(\mathbf{x})$ and the feature subspace is computed as follows:
\begin{equation}
\label{eq:theta}
\theta(\mathbf{x})=\arccos\frac{||\mathbf{R}^\mathrm{T}\overline{\mathbf{g}}(\mathbf{x})||_2}{||\overline{\mathbf{g}}(\mathbf{x})||_2}
\end{equation}
% From this definition, $\theta(\mathbf{x})\in\left[0,\frac{\pi}{2}\right]$. A smaller $\theta(\mathbf{x})$ implies that the component of the feature in the subspace is larger and closer to the origin, resulting in a higher likelihood of being an ID sample.

% \subsection{Enhancing PA by Entropy}
% Existing literature suggests that OOD detection requires the integration of multiple aspects, including features, logits, and softmax information \cite{b6}. In this work, we combine the PA with the widely used softmax-based entropy method from the existing references \cite{b30, b31}.

% $H(\mathbf{x})$ characterizes the concentration of output probabilities for sample $\mathbf{x}$, with ID samples expected to have lower $H(\mathbf{x})$ and OOD samples expected to have higher $H(\mathbf{x})$.

\noindent \textit{\textbf{The integration with Entropy.}} Note that $\mathbf{y} = \mathbf{W}^T\mathbf{g}(\mathbf{x})+\mathbf{b}$, we normalize $\mathbf{y}$ into probabilities using softmax: $p_i=e^{y_i}/\sum_{j=1}^Ce^{y_j}$, where $y_j$ represents the $j$-th component of $\mathbf{y}$. The softmax entropy of sample $\mathbf{x}$ is
\begin{equation}
\label{H}
    H(\mathbf{x})=-\sum_{i=1}^Cp_i\ln p_i
\end{equation}
Integrating PA and Entropy through simple addition does not effectively harness their respective advantages. Therefore, we introduce a training set-dependent scaling factor $\beta$ to strike a balance between PA and Entropy, calculated by 

\begin{equation}
    \beta=\frac{\max_{\mathbf{x}_s\in \mathbb{X}_s}H(\mathbf{x}_s)}{\min_{\mathbf{x}_s\in \mathbb{X}_s}\theta(\mathbf{x}_s)} 
\end{equation} 
where $\mathbb{X}_s$ is the sampled subset of the training set. Now we present the complete EPA score by weighting and adding PA and Entropy, given by
\begin{equation}
    \text{EPA}(\mathbf{x})=\beta\theta(\mathbf{x})+H(\mathbf{x})
\end{equation}

% Existing literature suggests that OOD detection requires a combination of various aspects of features, logits, and softmax [6]. Thus, we first compute a scaling factor $\beta=\frac{\sum_{\mathbf{x}_s\in\mathbb{X}_s}\max_i \mathbf{f}(\mathbf{x}_s)_i}{\sum_{\mathbf{x}_s\in\mathbb{X}_s}\theta_s}$, where $\mathbf{g}(\mathbf{x}_s)_i$ denotes the $i$-th component of $\mathbf{g}(\mathbf{x}_s)$, and $\theta_s$ is the principal angle for $\mathbf{x}_s$ as defined in equation (\ref{eq:theta}). Next, we use $-\log\left(\left({1}/{\text{softmax}\left(\begin{matrix}\beta\theta\\ \mathbf{f}(\mathbf{x})\end{matrix}\right)_1}\right)-1\right)$ as the complete OOD score. Here, $\text{softmax}\left(\begin{matrix}\beta\theta\\ \mathbf{f}(\mathbf{x})\end{matrix}\right)_1=\frac{e^{\beta\theta}}{e^{\beta\theta}+\sum_{i=1}^Ce^{\mathbf{f}(\mathbf{x})_i}}$ represents the first element of the softmax result. Expanding the above equation, we obtain the complete OOD score as follows:
% \begin{equation}
% \begin{aligned}
%     \text{EPA}(\mathbf{x})=&\beta\theta-\log \sum_{i=1}^C e^{\mathbf{f}(\mathbf{x})_i}\\
%     =&\beta\arccos\frac{||\mathbf{R}^\mathrm{T}\mathbf{g}^\prime (\mathbf{x})||_2}{||\mathbf{g}^\prime(\mathbf{x})||_2}-\log \sum_{i=1}^C e^{\mathbf{f}(\mathbf{x})_i}
% \end{aligned}
% \end{equation}

The specific implementation of EPA score is shown in Algorithm \ref{alg:spa}. A larger $\text{EPA}(\mathbf{x})$ implies a higher likelihood of being an OOD sample, while a smaller value suggests a higher likelihood of being an ID sample.

% \textbf{Comparison to Related Work}:  We note that \cite{b6} also uses $-\left(\mathbf{W}^\mathrm{T}\right)^{\dagger}\mathbf{b}$ to bias the features. Based on our novel analysis from $\mathcal{NC}$, we conclude that biasing the features with $-\left(\mathbf{W}^\mathrm{T}\right)^{\dagger}\mathbf{b}$ could further increase the component of ID features in the subspace, making it more advantageous for us to use the PA to distinguish between ID and OOD samples.

\section{Experiments}
\label{sec:typestyle}

\begin{figure}[t]
\vspace{-1em}
\begin{minipage}[b]{1\linewidth}
  \centering
  \centerline{\includegraphics[width=7.8cm]{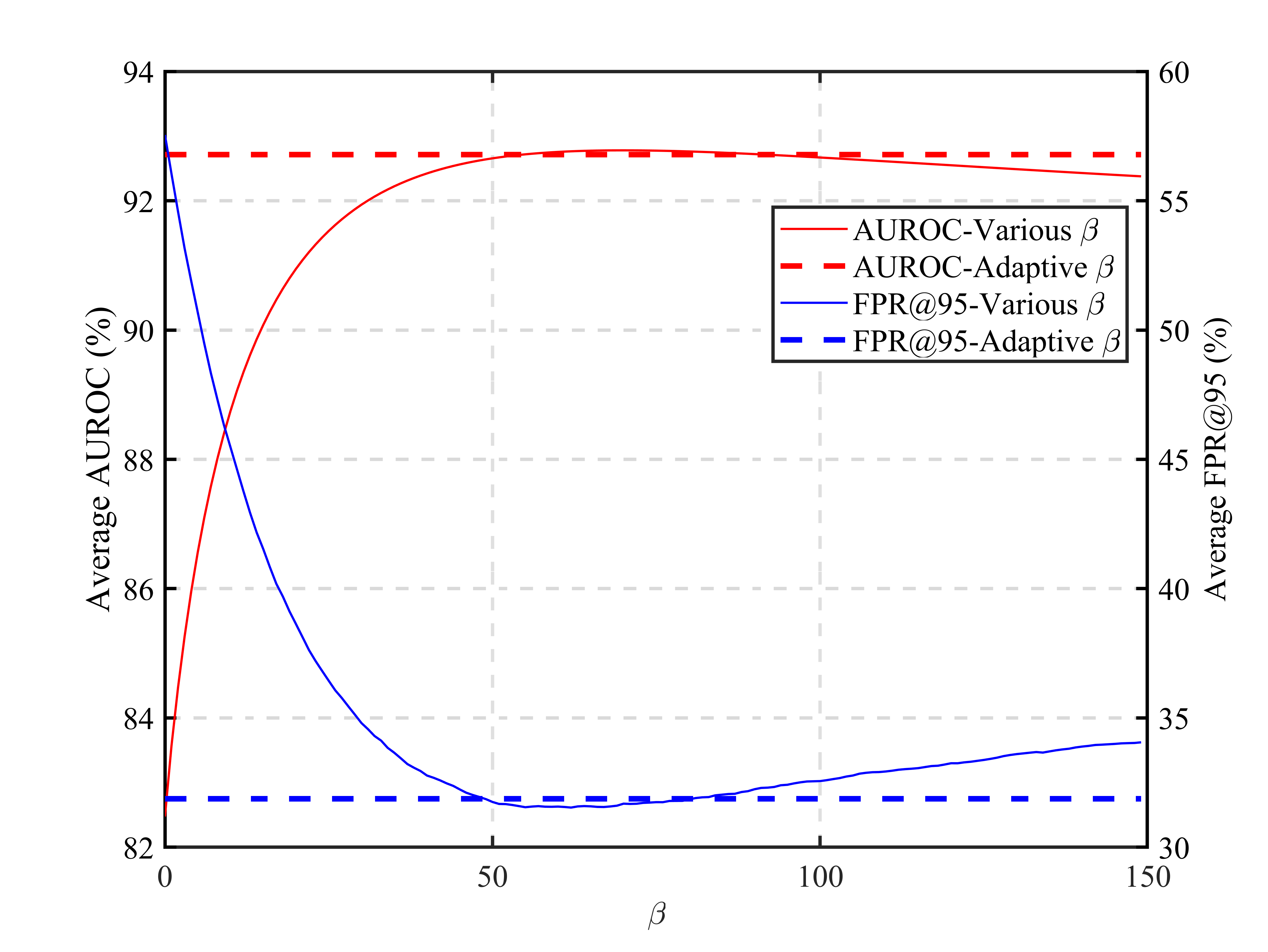}}
  \vspace{-1em}
\end{minipage}
\caption{Average AUROC\% and FPR@95\% for various $\beta$. The solid lines represent the performance of various $\beta$ range in $[0, 150]$. The dashed lines represent the performance using the adaptive $\beta$ in Section 3.2.}
\vspace{-1em}
\label{fig:2}
\end{figure}

\subsection{Experimental setup}

We use the validation set of ImageNet-1k as ID dataset and employ three different classification models: the CNN-based RepVGG \cite{b17}, ResNet-50d \cite{b18}, and the transformer-based Swin Transformer \cite{b19}, with their weights obtained from the timm repo\footnotemark{}\footnotetext{https://github.com/huggingface/pytorch-image-models}. We test on five different OOD datasets: Openimage-O \cite{b6}, Texture \cite{b21}, iNaturalist \cite{b22}, ImageNet-O \cite{b23}, and ImageNet-c \cite{b24}. For experiments on ImageNet-c, we use two subsets, ``blur" and ``digital" as the OOD dataset, encompassing a total of eight different corruptions and five corruption levels. AUROC and FPR@95 are used to evaluate the OOD detection methods. All results are reported in percentage.

We compare the performance of our proposed EPA score with various post-hoc methods, including classic OOD scores such as MSP \cite{b2}, Energy \cite{b3}, and MaxLogit \cite{b4}, as well as recent SOTA methods like ReAct \cite{b9}, ViM \cite{b6}, KNN \cite{b10}, DML \cite{b11}, and GEN \cite{b12}. We pre-sample 200,000 images from the ImageNet training set for all methods requiring training samples. For KNN, we set the number of nearest neighbors, $k$, to 150, which is tuned within the range [100, 300]. $\lambda$ is set to 10 for DML. Regarding GEN, we report the performance of the best-performing method in \cite{b12}, GEN+Residual. For the EPA score proposed in this work, we set the subspace dimensionality D to 1000 for RepVGG and ResNet-50d, which is consistent with the total number of ID categories $C$. For Swin Transformer, we set $D=512$. 
% Despite our theoretical analysis in Section 3.1 suggesting $D=C$, we find that the EPA score still outperforms other methods on Swin Transformer.
\begin{table}[htb]
\vspace{-0.5em}
\caption{Ablation study across four datasets. $\theta$ and $H$ refer to PA and Entropy respectively. `w/o $o^\prime$' refers to replacing $o^\prime$ by global mean $\mu_g$.\\}
\setlength{\tabcolsep}{7pt}
\centering
\resizebox{0.48\textwidth}{!}{
% \usepackage{multirow}
\begin{tabular}{llcccc}
\hline
Dataset                               & Metrics           & \textbf{EPA}   & \textbf{$\theta$} & \textbf{$H$} & w/o $o^\prime$ \\ \hline
\multirow{2}{*}{\textbf{Openimage-O}} & AUROC$\uparrow$   & \textbf{93.56} & 89.77             & 88.03        & 91.70                   \\
                                      & FPR95$\downarrow$ & \textbf{31.19} & 44.74             & 50.70        & 38.68                   \\ \hline
\multirow{2}{*}{\textbf{Texture}}     & AUROC$\uparrow$   & \textbf{96.73} & 96.47             & 83.06        & 95.22                   \\
                                      & FPR95$\downarrow$ & \textbf{14.44} & 15.78             & 60.53        & 20.58                   \\ \hline
\multirow{2}{*}{\textbf{iNaturalist}} & AUROC$\uparrow$   & \textbf{94.99} & 90.69             & 90.98        & 91.81                   \\
                                      & FPR95$\downarrow$ & \textbf{26.63} & 43.15             & 41.15        & 39.84                   \\ \hline
\multirow{2}{*}{\textbf{ImageNet-O}}  & AUROC$\uparrow$   & 85.57          & \textbf{86.56}    & 67.83        & 82.68                   \\
                                      & FPR95$\downarrow$ & 55.23          & \textbf{53.67}    & 77.8         & 60.55                   \\ \hline
\multirow{2}{*}{\textbf{Average}}     & AUROC$\uparrow$   & \textbf{92.71} & 90.87             & 82.48        & 90.35                   \\
                                      & FPR95$\downarrow$ & \textbf{31.87} & 39.34             & 57.55        & 39.91                   \\ \hline
\end{tabular}
}
\label{tab:2}
\vspace{-1em}
\end{table}
\subsection{Results and Analysis}
\begin{table*}[t]
\setlength{\tabcolsep}{6.2pt}
\centering
\caption{OOD detection performance of EPA and eight other algorithms on RepVGG, ResNet-50d, and Swin Transformer. OOD datasets include Openimage-O, Texture, iNaturalist, and ImageNet-O. AUROC\% and FPR@95\% metrics are reported. EPA demonstrates excellent performance across different models and OOD datasets.\\}
\resizebox{0.95\textwidth}{!}{
\begin{tabular}{llccccc}
\hline
\rowcolor[HTML]{FFFFFF} 
\cellcolor[HTML]{FFFFFF}                             & \multicolumn{1}{c}{\cellcolor[HTML]{FFFFFF}{\color[HTML]{333333} }}                                  & {\color[HTML]{333333} \textbf{Openimage-O}}             & {\color[HTML]{333333} \textbf{Texture}}                                         & {\color[HTML]{333333} \textbf{iNaturalist}}             & {\color[HTML]{333333} \textbf{ImageNet-O}}              & {\color[HTML]{333333} \textbf{Average}}                 \\
\rowcolor[HTML]{FFFFFF} 
\multirow{-2}{*}{\cellcolor[HTML]{FFFFFF}Model}      & \multicolumn{1}{c}{\multirow{-2}{*}{\cellcolor[HTML]{FFFFFF}{\color[HTML]{333333} \textbf{Method}}}} & {\color[HTML]{333333} AUROC$\uparrow$FPR95$\downarrow$} & \cellcolor[HTML]{FFFFFF}{\color[HTML]{333333} AUROC$\uparrow$FPR95$\downarrow$} & {\color[HTML]{333333} AUROC$\uparrow$FPR95$\downarrow$} & {\color[HTML]{333333} AUROC$\uparrow$FPR95$\downarrow$} & {\color[HTML]{333333} AUROC$\uparrow$FPR95$\downarrow$} \\ \hline
\rowcolor[HTML]{FFFFFF} 
\cellcolor[HTML]{FFFFFF}                             & {\color[HTML]{333333} MSP \cite{b2}}                                                                           & {\color[HTML]{333333} 85.06 63.36}                      & {\color[HTML]{333333} 78.58 72.62}                                              & {\color[HTML]{333333} 87.11 54.93}                      & {\color[HTML]{333333} 61.65 91.30}                      & {\color[HTML]{333333} 78.10 70.55}                      \\
\rowcolor[HTML]{FFFFFF} 
\cellcolor[HTML]{FFFFFF}                             & {\color[HTML]{333333} Energy \cite{b3}}                                                                        & {\color[HTML]{333333} 83.64 69.92}                      & {\color[HTML]{333333} 74.53 82.97}                                              & {\color[HTML]{333333} 83.92 75.31}                      & {\color[HTML]{333333} 63.36 87.75}                      & {\color[HTML]{333333} 76.36 78.99}                      \\
\rowcolor[HTML]{FFFFFF} 
\cellcolor[HTML]{FFFFFF}                             & {\color[HTML]{333333} MaxLogit \cite{b4}}                                                                      & {\color[HTML]{333333} 84.81 65.04}                      & {\color[HTML]{333333} 76.33 76.86}                                              & {\color[HTML]{333333} 86.22 62.20}                      & {\color[HTML]{333333} 62.87 89.90}                      & {\color[HTML]{333333} 77.56 73.50}                      \\
\rowcolor[HTML]{FFFFFF} 
\cellcolor[HTML]{FFFFFF}                             & {\color[HTML]{333333} ReAct \cite{b9}}                                                                         & {\color[HTML]{333333} 46.08 99.65}                      & {\color[HTML]{333333} 54.56 97.66}                                              & {\color[HTML]{333333} 47.18 99.88}                      & {\color[HTML]{333333} 48.76 98.65}                      & {\color[HTML]{333333} 49.14 98.96}                      \\
\rowcolor[HTML]{FFFFFF} 
\cellcolor[HTML]{FFFFFF}                             & {\color[HTML]{333333} ViM \cite{b6}}                                                                           & {\color[HTML]{333333}  {89.27}  {52.40}}                      & {\color[HTML]{333333} 93.69 23.76}                                              & {\color[HTML]{333333} 91.35 46.79}                      & {\color[HTML]{333333}   {76.93} 79.05}                      & {\color[HTML]{333333} 87.81 50.50}                      \\
\rowcolor[HTML]{FFFFFF} 
\cellcolor[HTML]{FFFFFF}                             & {\color[HTML]{333333} KNN \cite{b10}}                                                                           & {\color[HTML]{333333} 88.33 54.54}                      & {\color[HTML]{333333}   {94.98}   {14.69}}                                              & {\color[HTML]{333333} 89.33 62.57}                      & {\color[HTML]{333333} \textbf{83.38 62.45}}             & {\color[HTML]{333333}   {89.01}   {48.56}}                      \\
\rowcolor[HTML]{FFFFFF} 
\cellcolor[HTML]{FFFFFF}                             & {\color[HTML]{333333} DML \cite{b11}}                                                                           & {\color[HTML]{333333} 84.19 72.53}                      & {\color[HTML]{333333} 84.07 70.45}                                              & {\color[HTML]{333333} 85.61 73.33}                      & {\color[HTML]{333333} 71.26 92.50}                      & {\color[HTML]{333333} 81.28 77.20}                      \\
\rowcolor[HTML]{FFFFFF} 
\cellcolor[HTML]{FFFFFF}                             & {\color[HTML]{333333} GEN+Res \cite{b12}}                                                                       & {\color[HTML]{333333} 88.99 53.84}                      & {\color[HTML]{333333} 92.74 27.87}                                              & {\color[HTML]{333333}   {92.17}   {42.74}}                      & {\color[HTML]{333333} 76.11 82.10}                      & {\color[HTML]{333333} 87.50 51.64}                      \\
\rowcolor[HTML]{FFFFFF} 
\multirow{-9}{*}{\cellcolor[HTML]{FFFFFF}RepVGG}     & {\color[HTML]{333333} \textbf{EPA (ours)}}                                                            & {\color[HTML]{333333} \textbf{91.16 38.84}}             & {\color[HTML]{333333} \textbf{97.19 9.65}}                                     & {\color[HTML]{333333} \textbf{93.21 34.63}}             & {\color[HTML]{333333} 82.11  63.50}                      & {\color[HTML]{333333} \textbf{90.92 36.66}}             \\ \hline
\rowcolor[HTML]{FFFFFF} 
\cellcolor[HTML]{FFFFFF}                             & {\color[HTML]{333333} MSP \cite{b2}}                                                                           & {\color[HTML]{333333} 84.50 63.53}                      & {\color[HTML]{333333} 82.75 64.40}                                              & {\color[HTML]{333333} 88.58 50.05}                      & {\color[HTML]{333333} 56.13 93.85}                      & {\color[HTML]{333333} 77.99 67.96}                      \\
\rowcolor[HTML]{FFFFFF} 
\cellcolor[HTML]{FFFFFF}                             & {\color[HTML]{333333} Energy \cite{b3}}                                                                        & {\color[HTML]{333333} 75.95 76.83}                      & {\color[HTML]{333333} 73.93 75.31}                                              & {\color[HTML]{333333} 80.50 71.32}                      & {\color[HTML]{333333} 53.95 90.10}                      & {\color[HTML]{333333} 71.08 78.39}                      \\
\rowcolor[HTML]{FFFFFF} 
\cellcolor[HTML]{FFFFFF}                             & {\color[HTML]{333333} MaxLogit \cite{b4}}                                                                      & {\color[HTML]{333333} 81.50 65.50}                      & {\color[HTML]{333333} 79.25 66.20}                                              & {\color[HTML]{333333} 86.42 53.00}                      & {\color[HTML]{333333} 54.39 92.65}                      & {\color[HTML]{333333} 75.39 69.34}                      \\
\rowcolor[HTML]{FFFFFF} 
\cellcolor[HTML]{FFFFFF}                             & {\color[HTML]{333333} ReAct \cite{b9}}                                                                         & {\color[HTML]{333333} 85.30 60.79}                      & {\color[HTML]{333333} 91.12 39.26}                                              & {\color[HTML]{333333} 87.27 56.03}                      & {\color[HTML]{333333} 68.02 78.45}                      & {\color[HTML]{333333} 82.93 58.63}                      \\
\rowcolor[HTML]{FFFFFF} 
\cellcolor[HTML]{FFFFFF}                             & {\color[HTML]{333333} ViM \cite{b6}}                                                                           & {\color[HTML]{333333}   {90.76}   {50.45}}                      & {\color[HTML]{333333} 95.84   {20.58}}                                              & {\color[HTML]{333333} 89.26 64.59}                      & {\color[HTML]{333333} 81.02 74.80}                      & {\color[HTML]{333333} 89.22 52.61}                      \\
\rowcolor[HTML]{FFFFFF} 
\cellcolor[HTML]{FFFFFF}                             & {\color[HTML]{333333} KNN \cite{b10}}                                                                           & {\color[HTML]{333333} 80.40 76.91}                      & {\color[HTML]{333333}   {95.95} 20.68}                                              & {\color[HTML]{333333} 79.61 84.50}                      & {\color[HTML]{333333} \textbf{84.05} \textbf{65.75}}             & {\color[HTML]{333333} 85.00 61.96}                      \\
\rowcolor[HTML]{FFFFFF} 
\cellcolor[HTML]{FFFFFF}                             & {\color[HTML]{333333} DML \cite{b11}}                                                                           & {\color[HTML]{333333} 86.21 67.58}                      & {\color[HTML]{333333} 88.11 57.98}                                              & {\color[HTML]{333333} 87.15 66.68}                      & {\color[HTML]{333333} 71.96 91.80}                      & {\color[HTML]{333333} 83.36 71.01}                      \\
\rowcolor[HTML]{FFFFFF} 
\cellcolor[HTML]{FFFFFF}                             & {\color[HTML]{333333} GEN+Res \cite{b12}}                                                                       & {\color[HTML]{333333} 90.17 53.34}                      & {\color[HTML]{333333} 95.24 23.47}                                              & {\color[HTML]{333333} 90.65 58.19}             & {\color[HTML]{333333} 80.23 78.40}                      & {\color[HTML]{333333} 89.07 53.35}                      \\
\rowcolor[HTML]{FFFFFF} 
\multirow{-9}{*}{\cellcolor[HTML]{FFFFFF}ResNet-50d} & {\color[HTML]{333333} \textbf{EPA (ours)}}                                                            & {\color[HTML]{333333} \textbf{92.07 40.47}}             & {\color[HTML]{333333} \textbf{97.57 11.92}}                                     & {\color[HTML]{333333}  \textbf{92.40 42.43}}                      & {\color[HTML]{333333} 81.97 66.55}                      & {\color[HTML]{333333} \textbf{91.00 40.34}}             \\ \hline
\rowcolor[HTML]{FFFFFF} 
\cellcolor[HTML]{FFFFFF}                             & {\color[HTML]{333333} MSP \cite{b2}}                                                                           & {\color[HTML]{333333} 91.35 34.96}                      & {\color[HTML]{333333} 85.21 51.90}                                              & {\color[HTML]{333333} 94.76 23.19}                      & {\color[HTML]{333333} 78.97 63.70}                      & {\color[HTML]{333333} 87.57 43.44}                      \\
\rowcolor[HTML]{FFFFFF} 
\cellcolor[HTML]{FFFFFF}                             & {\color[HTML]{333333} Energy \cite{b3}}                                                                        & {\color[HTML]{333333} 90.93 27.58}                      & {\color[HTML]{333333} 82.62 51.57}                                              & {\color[HTML]{333333} 95.22 15.47}                      & {\color[HTML]{333333} 82.29 45.70}                      & {\color[HTML]{333333} 87.77 35.08}                      \\
\rowcolor[HTML]{FFFFFF} 
\cellcolor[HTML]{FFFFFF}                             & {\color[HTML]{333333} MaxLogit \cite{b4}}                                                                      & {\color[HTML]{333333} 91.91 26.79}                      & {\color[HTML]{333333} 84.67 47.42}                                              & {\color[HTML]{333333} 95.72 15.41}                      & {\color[HTML]{333333} 81.28 51.50}                      & {\color[HTML]{333333} 88.40 35.28}                      \\
\rowcolor[HTML]{FFFFFF} 
\cellcolor[HTML]{FFFFFF}                             & {\color[HTML]{333333} ReAct \cite{b9}}                                                                         & {\color[HTML]{333333} 93.58 23.07}                      & {\color[HTML]{333333} 85.51 47.91}                                              & {\color[HTML]{333333} 97.51 \text{ }\text{ }9.98}                       & {\color[HTML]{333333} 84.09 44.50}                      & {\color[HTML]{333333} 90.17 31.36}                      \\
\rowcolor[HTML]{FFFFFF} 
\cellcolor[HTML]{FFFFFF}                             & {\color[HTML]{333333} ViM \cite{b6}}                                                                           & {\color[HTML]{333333} 96.04 23.88}                      & {\color[HTML]{333333} 92.34 38.49}                                              & {\color[HTML]{333333} 99.28\text{ } \text{ }\textbf{2.60}}                       & {\color[HTML]{333333} 88.78 59.20}                      & {\color[HTML]{333333} 94.11 31.04}                      \\
\rowcolor[HTML]{FFFFFF} 
\cellcolor[HTML]{FFFFFF}                             & {\color[HTML]{333333} KNN \cite{b10}}                                                                           & {\color[HTML]{333333} 92.86 41.19}                      & {\color[HTML]{333333} 91.13 38.00}                                              & {\color[HTML]{333333} 95.46 27.51}                      & {\color[HTML]{333333} 86.79 58.40}                      & {\color[HTML]{333333} 91.56 41.28}                      \\
\rowcolor[HTML]{FFFFFF} 
\cellcolor[HTML]{FFFFFF}                             & {\color[HTML]{333333} DML \cite{b11}}                                                                           & {\color[HTML]{333333} 95.09 25.39}                      & {\color[HTML]{333333} 89.21 45.97}                                              & {\color[HTML]{333333} 97.75 11.33}                      & {\color[HTML]{333333} 86.20 57.25}                      & {\color[HTML]{333333} 92.06 34.99}                      \\
\rowcolor[HTML]{FFFFFF} 
\cellcolor[HTML]{FFFFFF}                             & {\color[HTML]{333333} GEN+Res \cite{b12}}                                                                       & {\color[HTML]{333333} 95.70 25.42}                      & {\color[HTML]{333333} 92.18 38.33}                                              & {\color[HTML]{333333} 99.12 \text{ }\text{ }3.16}                       & {\color[HTML]{333333} 88.10 62.05}                      & {\color[HTML]{333333} 93.78 32.24}                      \\
\rowcolor[HTML]{FFFFFF} 
\multirow{-9}{*}{\cellcolor[HTML]{FFFFFF}Swin}       & {\color[HTML]{333333} \textbf{EPA (ours)}}                                                            & {\color[HTML]{333333} \textbf{97.44 14.26}}             & {\color[HTML]{333333} \textbf{95.44 21.76}}                                     & {\color[HTML]{333333} \textbf{99.36}
\text{ } 2.84}              & {\color[HTML]{333333} \textbf{92.64 35.65}}             & {\color[HTML]{333333} \textbf{96.22 18.63}}             \\ \hline
\end{tabular}
}
\vspace{-1em}
\label{tab:1}
\end{table*}
% \usepackage{multirow}
\noindent \textit{\textbf{Ablation study.}}
Table \ref{tab:2} showcases the results of ablation experiments. We compare the performance of the proposed EPA score, the PA $\theta$, and the Entropy $H$. We also compare the effect of biasing the feature by $o^\prime$ and by global mean $\mu_g$ (referred to as w/o $o^\prime$). All values are averaged across three different models. The results demonstrate that PA is a discriminative OOD score, and the enhancement of PA by Entropy in this paper leads to a significant improvement in performance. Moreover, replacing $\mu_g$ by $o^\prime$ to bias the features results in a 2.36\% improvement in AUROC and an 8.04\% reduction in FPR@95.
\begin{figure}[ht]
% \vspace{-1em}
\begin{minipage}[b]{1\linewidth}
  \centering
  \centerline{\includegraphics[width=7cm]{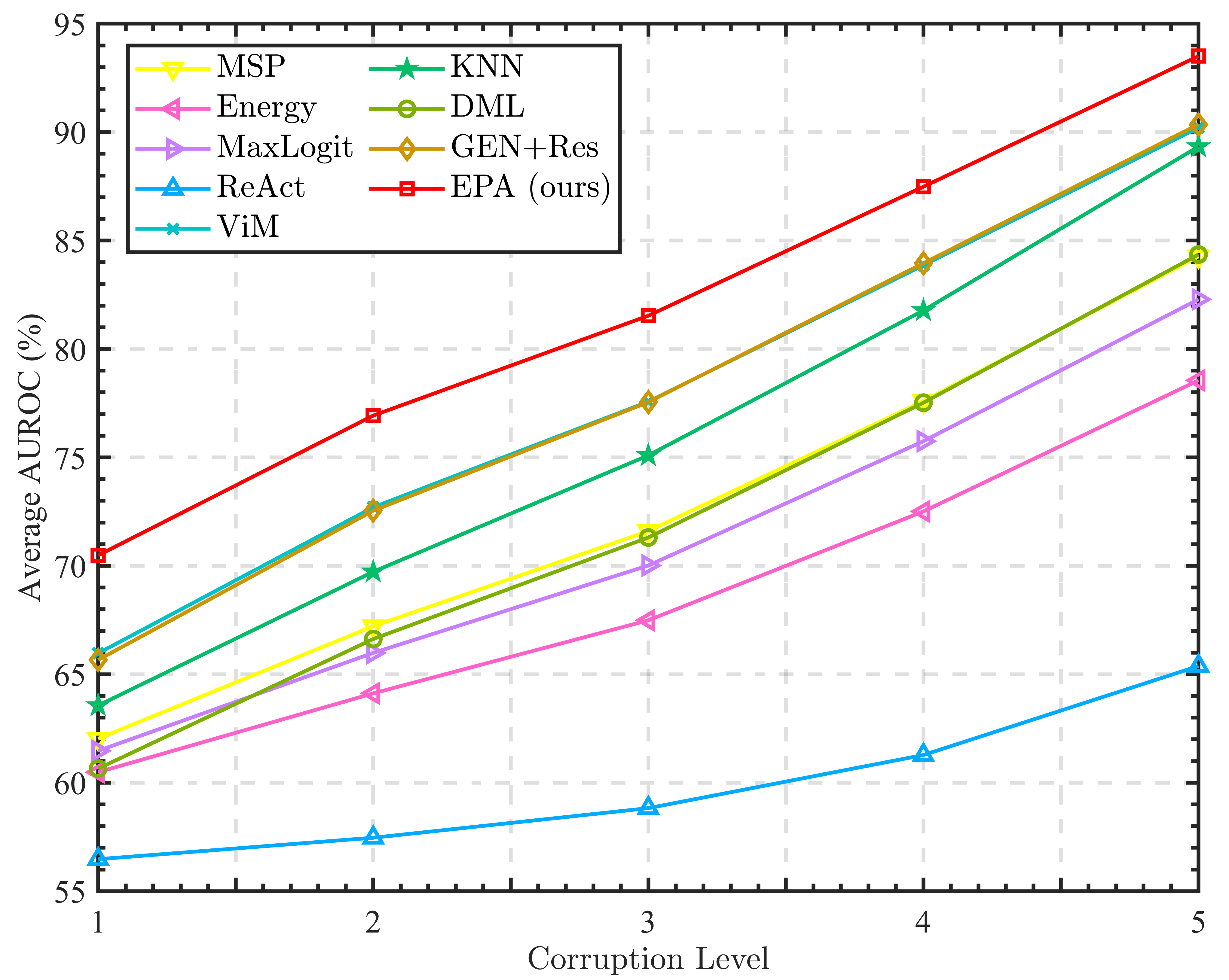}}
  \vspace{-1em}
\end{minipage}
\caption{Average AUROC\% of different algorithms on ImageNet-c, averaged across RepVGG, ResNet-50d, and Swin Transformer. The horizontal axis represents different corruption levels. EPA demonstrates excellent performance across different corruption levels.}
\vspace{-1em}
\label{fig:1}
\end{figure}

We further validate the effectiveness of the selection of $\beta$ outlined in Section 3.2. The solid line in Figure \ref{fig:2} illustrates the average AUROC and FPR@95 across three models and four datasets (Openimage-O, Texture, iNaturalist, ImageNet-O) for various $\beta$ values, while the dashed line represents the average performance using the $\beta$ in Section 3.2. The results demonstrate that the proposed selection of $\beta$, as an adaptive criterion that requires no manual tuning, achieves competitive performance.

\noindent \textit{\textbf{Comparison with other methods.}}
Table 2 presents the performance of EPA and other methods on three different models (RepVGG, ResNet-50d, Swin Transformer) and four OOD datasets (Openimage-O, Texture, iNaturalist, ImageNet-O). The last column displays the average AUROC and FPR@95 values across different OOD datasets. EPA exhibits highly competitive performance across various models and OOD datasets, and consistently achieves the highest average AUROC and the lowest average FPR@95 across three models. This demonstrates the robustness and adaptability of EPA to different OOD datasets and various deep neural network architectures, including CNN-based and Transformer-based models.

We also evaluate EPA's performance in detecting distribution shifts. Figure \ref{fig:1} displays the average AUROC across RepVGG, ResNet-50d, and Swin Transformer of nine methods when using ImageNet-c's ``blur" and ``digital" subsets as OOD dataset. The corruptions are categorized into five levels, with level 1 having minimal corruption (indicating OOD data is closer to ID data and OOD detection is the toughest) and level 5 having the highest corruption (indicating the easiest OOD detection). Figure \ref{fig:1} showcases that EPA consistently achieves the highest AUROC across different corruption levels, confirming the robustness of our EPA score.

% \usepackage{multirow}
% \usepackage{multirow}
% \usepackage{multirow}

\section{Conclusion}
In this paper, we propose EPA, a novel method for OOD detection based on new observations drawn from Neural Collapse. We conduct a comprehensive ablation study and compare the proposed method with various algorithms on multiple models and OOD datasets, confirming the outstanding performance and robustness of our EPA method. Future work may delve into the distribution of OOD samples and explore further improvements to the $\mathcal{NC}$-inspired methods.
\newpage
% Out-of-distribution (OOD) detection plays a crucial role in ensuring the security of neural networks. Recent studies have leveraged the fact that In-distribution (ID) samples form a subspace in the feature space, achieving state-of-the-art (SOTA) performance. However, these methods overlook the fine-grained distribution of ID features within the feature subspace. Motivated by the $\mathcal{NC}$, we observe that ID features tend to deviate from the global feature mean and concentrate around the feature mean of their corresponding ID class. Building upon these two observations, we introduce a novel $\mathcal{NC}$-inspired OOD scoring function, named Entropy-enhanced Principal Angle (EPA). This method quantifies the proximity of a test feature to the global ID subspace by calculating the principal angle between the feature and the global ID subspace. Furthermore, it measures the proximity of the test sample to the nearest ID class feature mean by computing the softmax entropy. We experimentally compare EPA with various SOTA approaches, validating its superior performance and robustness across different network architectures and OOD datasets.
\vfill\pagebreak
\bibliographystyle{IEEEbib}
\bibliography{strings,refs}

\end{document}